\documentclass[letterpaper, 10 pt, conference]{ieeeconf}  
\IEEEoverridecommandlockouts                              
\usepackage{graphics} 
\usepackage{epsfig} 
\usepackage{amsmath} 
\usepackage{amssymb}  
\usepackage{xcolor}
\usepackage{svg}
\usepackage{hyperref}

\usepackage{multirow}
\usepackage[normalem]{ulem}
\useunder{\uline}{\ul}{}

\title{HADRON: Human-friendly Control and Artificial Intelligence for Military Drone Operations}

\author{Ana M. Casado Fauli$^{1}$, Mario Malizia$^{1}$, Ken Hasselmann$^{1}$, Emile Le Flécher$^{1}$, Geert De Cubber $^{1}$,  Ben Lauwens$^{2}$  
\thanks{$^{1}$Dept. of Mechanics, Royal Military Academy of Brussels, Belgium.}
\thanks{$^{1}$Dept. of Mathematics, Royal Military Academy of Brussels, Belgium.}
}

\begin{document}
\maketitle


\begin{abstract}
As drones are getting more and more entangled in our society, more untrained users require the capability to operate them. This scenario is to be achieved through the development of artificial intelligence capabilities assisting the human operator in controlling the Unmanned Aerial System (UAS) and processing the sensor data, thereby alleviating the need for extensive operator training. This paper presents the HADRON project that seeks to develop and test multiple novel technologies to enable human-friendly control of drone swarms. This project is divided into three main parts. The first part consists of the integration of different technologies for the intuitive control of drones, focusing on novice or inexperienced pilots and operators. The second part focuses on the development of a multi-drone system that will be controlled from a command and control station, in which an expert pilot can supervise the operations of the multiple drones. The third part of the project will focus on reducing the cognitive load on human operators, whether they are novice or expert pilots. For this, we will develop AI tools that will assist drone operators with semi-automated real-time data processing.
\end{abstract}


\section{INTRODUCTION} \label{sec-intro}
The use of Unmanned Aerial Systems (UASs) or drones has been exploding in recent years~\cite{Daponte_2019, drones4040064, POLJAK2020425, banu2016use}. This poses challenges related to the training of the personnel able to operate these tools. In this paper, we will tackle the multi-modal control problem of collaborative drones in a military use case. For future military operations, it can be envisioned that many soldiers/pilots will possess the capability to operate drones, with drones being regarded as one among numerous tools available to individual human operators\cite{AYAMGA2021120677}.

However, in order to arrive at such a situation, there are still multiple obstacles to be tackled. A significant challenge lies in the complexity of both controlling and interpreting data from drones, rendering it impractical for personnel without extensive training to handle. Nevertheless, Defense training structures do not allow providing extensive training programs to work as drone pilots for intricate operations \cite{drone_training}. This means that, in the future, drones will also be placed in the hands of quite \textit{novice} users, such that the \textit{expert} users can concentrate on more complex operations. Catering to these different skill-levels of drone pilots, the level of autonomy for the drones should also be varied and optimized to the use case. 

The HADRON project plans to render the use of drones easier for the operators while tackling the challenge of scalability concerning drone usage in military settings. This is achieved through the development of artificial intelligence capabilities providing an optimal level of assistance to the human operator in controlling the UAS and in processing the sensor data, thereby alleviating the need for extensive operator training. 


The HADRON project considers three levels of autonomy. The level focused on novice users requiring a high degree of assistance even for fairly simple operations, e.g. through the use of technologies for intuitive drone control. The next level concerns expert pilots and aims at giving them the ability to control multi-drone systems from a central station. A final level aims at drastically reducing the cognitive load of the operators, by introducing artificial intelligence tools to automate real-time data processing.

This article presents the definition of the HADRON project, where different novel technologies will be developed and tested, to clarify what will be useful to the future of drone piloting. This paper presents the system architecture and discusses the choice of the different technologies that are used.


\section{STATE OF THE ART} \label{sec-context}
A better understanding of the control concepts for UASs~\cite{balta20203d} in the last two decades have rendered possible the development of advanced autopilot systems. These systems have notably reduced the challenges associated with controlling drones, thereby driving their widespread adoption. Commercial drone manufacturers have integrated diverse new control options in their products. Some examples include:


\begin{itemize}
    \item Visual remote radio control~\cite{barnhart2021introduction}, which implies visual line of sight. While reliable, it limits operational range and can make determining the orientation difficult at long distances.
    \item Screen-based remote control adds a screen to visual remote control. This allows for both visual line of sight and beyond visual line of sight operations, improving the awareness of the orientation of the drone. However, it still requires the full attention of the operator.
    \item First-person-view remote control~\cite{fpvdrones} simulates on-board flying for the operator and is popular in drone racing~\cite{9372809}. It requires full attention and isolates operators from their environment, posing potential hazards in demanding operations.
    \item Ground Control or Command and Control Stations~\cite{prisacariu2021considerations} provide information about the environment and drone status, reducing the cognitive load required for operation. 
\end{itemize}

In the research community, alternative control modalities have been proposed, such as gesture-based control~\cite{kim2022intuitive}, auditive control~\cite{computers9030075}, brain-computer interfaces~\cite{app112411876}, augmented reality~\cite{de2019explosive} and multimodal interfaces~\cite{lahouli2019pedestrian1}. These control modalities provide promising first results as a mean to assist non-expert drone operators with the possibility to control an UAS while focusing on other tasks. However, these modalities have not yet found their way to widespread adoption. 

Drone operations are becoming increasingly automated. Autonomous flight, where UASs can operate independently using advanced control algorithms and navigation systems, represents a leading edge in drone technology. This advancement allows UAVs to fly, navigate, and carry out tasks on their own, opening up exciting avenues for further research and development~\cite{telli2023comprehensive}. Due to this, expert operators may focus primarily on overseeing critical tasks such as take-off, landing, or managing unexpected issues or malfunctions. As such, it becomes possible for one expert user to control multiple drones, as shown in~\cite{lahouli2019pedestrian2}. 

The use of multi-agent systems has been a topic of interest in robotics, as the use of multiple robots exceeds the capabilities of a single device doing all the work. In the case of UASs, such systems can help to overcome problems such as battery usage or the coverage of large areas. Yet, having multiple drones working together simultaneously poses significant challenges due to the complexity of the communications and coordination among them, as the number of agents grows~\cite{leon2022multi}. This adds up to the synchronization of individual UAV actions while considering diverse mission objectives and real-time obstacles~\cite{lombard2020velocity}.


The operation of UASs is subject to high uncertainty, which requires a strong control and supervising system to detect malfunctions~\cite{yun2022cooperative}. At the same time, these types of operations can be sensitive to time constraints, and therefore the information should be presented comprehensibly and efficiently~\cite{balta2017integrated}, so the pilots do not waste time in understanding the data and concentrate on supervising the platforms and the mission. 

Processing data produced by UASs is a challenging task, let alone its subsequent interpretation by humans. For this, classification tools based on artificial intelligence show good performances. The importance of these tools is expected to increase with the advances of drones equipped with on-board GPU-units~\cite{9613590}. However, as these algorithms are application-specific, it is required to carefully design them for each specific use case. For example, in the case of automatic anomaly detection~\cite{de2019optimized}, the focus should rely only on the relevant data streams or application-specific drone data~\cite{balta2019semi, doroftei2019using}, extracting useful information from the drone. 


\section{SYSTEM OVERVIEW} \label{sec-system}
In view of the identified capability gaps in Section~\ref{sec-context}, three topics have been defined: the first one will develop human-friendly control of a drone for untrained users; the second one will develop multi-agent control of drones for expert users; lastly, the third topic will assist the two previous one by developing semi-automated data interpretation algorithms. 


\subsection{Human-Friendly Control of a Drone by Untrained Users} \label{subsec-untrained}
Firstly, the use cases selected will be explained. By understanding how the proposed technologies will be applied in specific situations, their potential impact and functionalities can be better understood. 

\begin{figure}[tbh]
    \centering
    \includegraphics[width=0.6\linewidth]{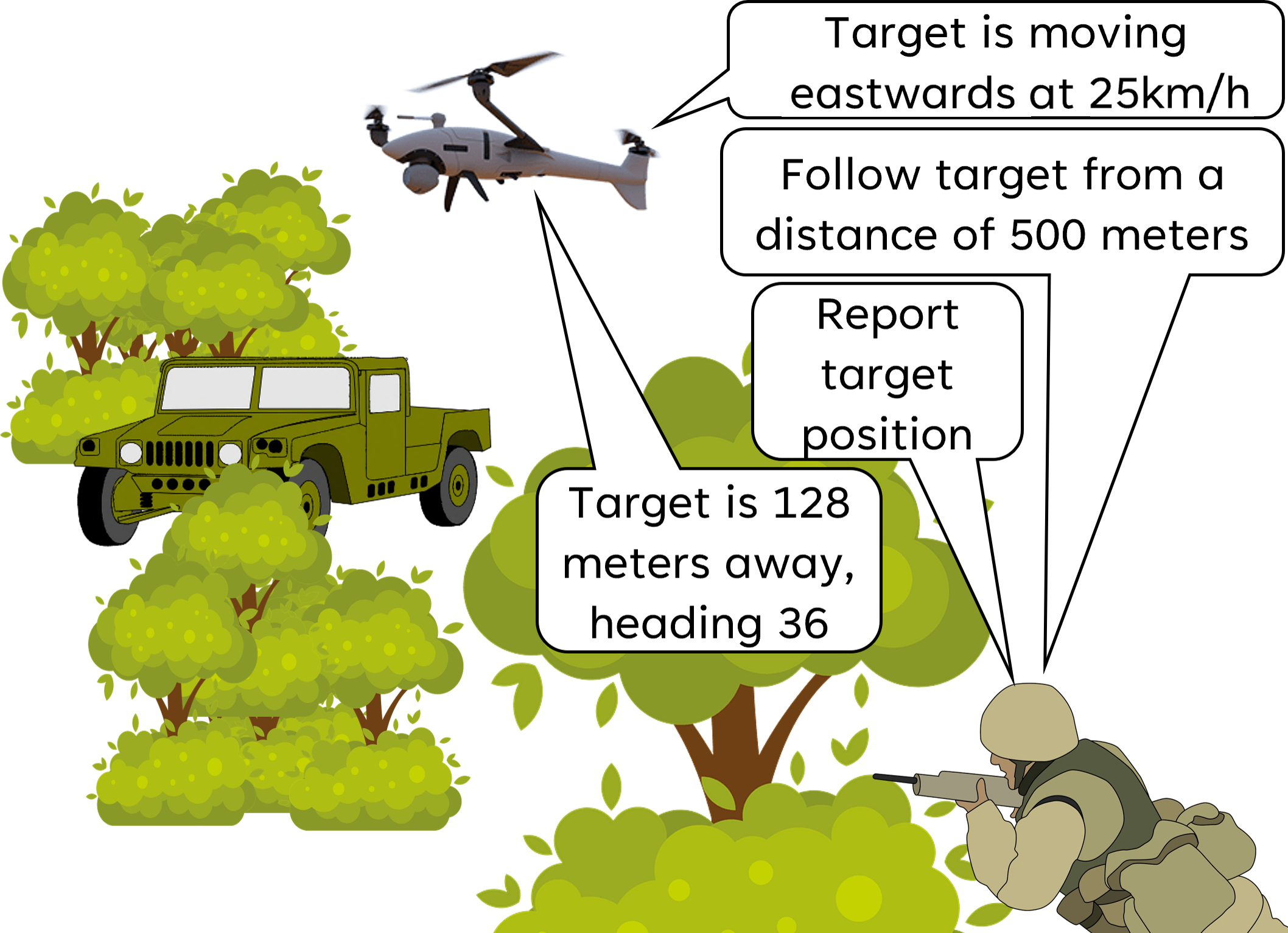}
    \caption{Depiction of a dismounted soldier operating a drone, using voice commands. The drone detects a target and informs the operator, via audio, so the operator can make decisions on the spot.}
    \label{fig:unmounted_soldier}
\end{figure}

Two scenarios are envisioned for this topic:
\begin{itemize}
    \item A dismounted soldier who is operating a drone. This means that they have to be aware of their surroundings as well as of the UAS they are managing, as portrayed in Figure~\ref{fig:unmounted_soldier}.
    \item Field or naval commander overseeing a mission, requiring direct control of a drone for tactical decisions.
\end{itemize}

This means that technologies will be tested both in static and dynamic environments. It is foreseen that not all the technologies will be integrated in the same way or even at all for both use cases, as the environments are completely different. In terms of research, though, it will be interesting to create a common framework for both use cases.

In the first place, a tablet control interface will be developed. This system will feature a user-friendly web application interface, enabling operators to access drone data in an easily understandable format, as well as having access to mission planning capabilities. This will include visualization of typical drone status information, such as odometry, battery status, GPS position, map visualization and trajectory designing involving the usage of waypoints.

One key aspect of this app will be its modularity, allowing operators to customize their interface according to their preferences. This emphasis on modularity comes from the understanding of the heterogeneous nature of the operational contexts, necessitating adaptable specifications. Another feature foreseen for the interface is its abstract nature, meaning it will be platform-agnostic. In this way, the application can be tailored to meet diverse operational requirements effectively. 

This tablet interface will be very useful for the commander use case, where the user is static. For the dismounted soldier, it is necessary to integrate more dynamic solutions. One of these technologies will be voice control, as depicted in Figure~\ref{fig:unmounted_soldier}. This will include noise-canceling microphones, speech recognition and text-to-speech functionalities. This integration aims to facilitate communication between the human operator and the UAS, even in noisy or challenging environments. Through the use of speech recognition technology, operators can verbally command the drone, freeing their hands for other tasks and reducing the cognitive load associated with manual input. The development of speech command recognition to keep operators informed during semi-autonomous drone-assisted tasks remains relatively limited in the literature~\cite{li2023multi}. Furthermore, the incorporation of text-to-speech capabilities will enable the drone to provide audible feedback or status updates to the operator, enhancing situational awareness and enabling decision-making in high-pressure scenarios.
Some examples of what will be implemented include drone commands, such as, take off and land, go to waypoint command, pre-programmed missions, drone and mission status updates and information about the environment.


The next prospected technology to be integrated is gesture-based control. In the literature, there are multiple instances of drone control where the hand gestures are recognized by cameras and then interpreted as drone commands~\cite{khaksar2023design, yoo2022motion}. Although results seem promising, it would be difficult to implement this kind of technology in dynamic environments. Therefore, the considered alternative for gesture-based control will integrate wearable motion sensors~\cite{kim2022intuitive,macchini2020hand}, such as haptic gloves. 
For this technology, only main drone commands are foreseen, for instance, take-off, land and roll, pitch, yaw, and thrust movements.

The final interface to be developed is the one based on augmented reality. This one will be interchangeable with the tablet control and will have the same essence of modularity and abstraction. The main goal is to send commands from this interface to the UAS and then portray the minimal and necessary information to the pilot. An interesting feature of this interface will rely on the capability of changing from the point of view of the operator to the first-person view of the drone, thus, enabling different ways of piloting depending on the user's preference. 

All defined technologies, including the tablet interface, voice commands, gesture-based control, and the augmented reality interface, will be designed with the capability of being integrated and interchanged, ensuring flexibility and modularity within the system. This modularity eases the customization of control interfaces according to user preferences and operational requirements, enhancing flexibility and usability across diverse scenarios.


\subsection{Multi-agent Control of Drones by Expert Users} \label{subsec-expert}

\begin{figure}[ht]
    \centering
    \includegraphics[width=0.7\linewidth]{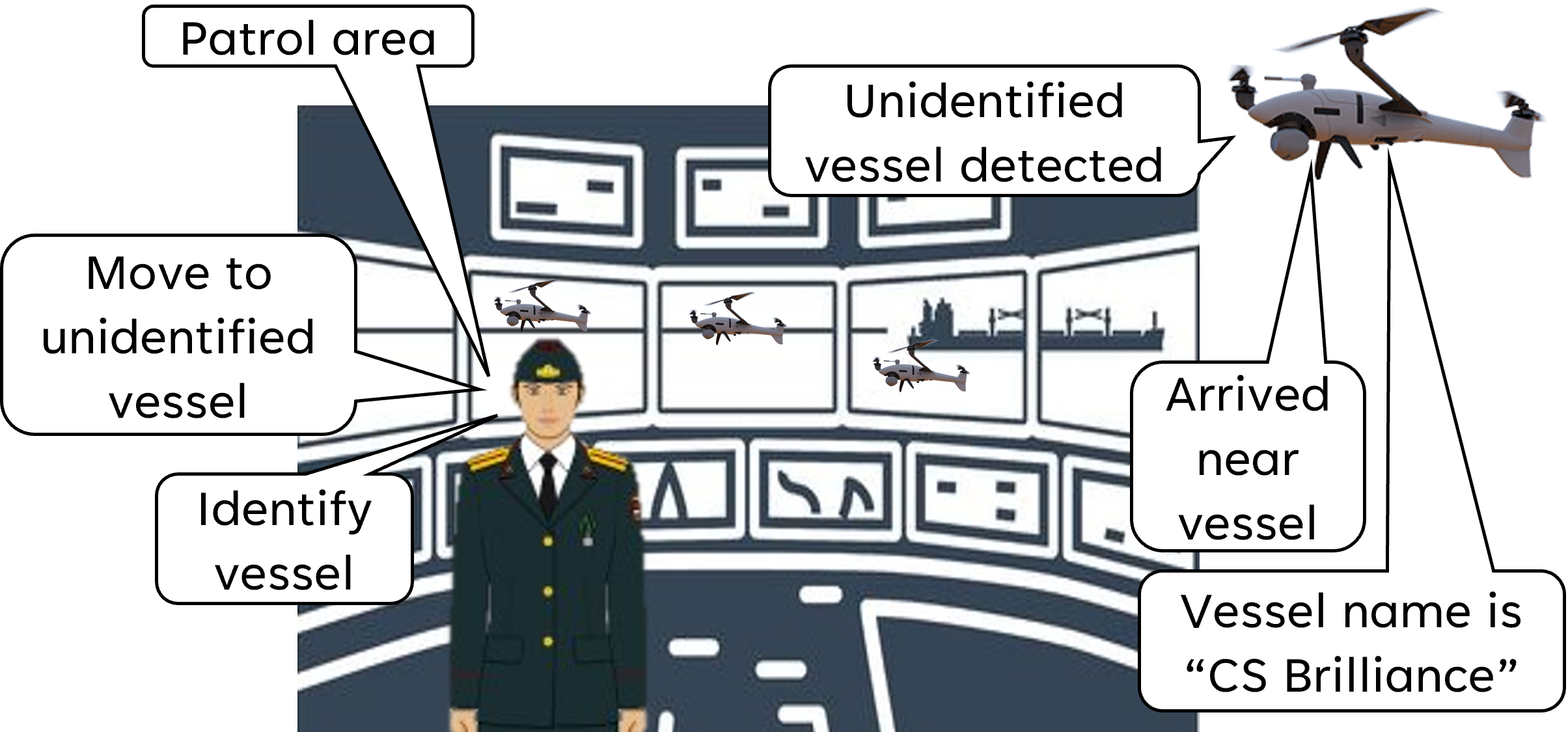}
    \caption{Depiction of a command officer operating a UAS patrol system, using voice commands. After a patrol mission has been deployed, if the drones detect an intruder, they identify it and inform the operator about it.}
    \label{fig:expert_soldier}
\end{figure}

As in Section~\ref{subsec-untrained}, first the use cases will be presented. Taking into account that an expert pilot is in charge of several UASs, the use cases portray the operations that the pilot will be doing instead of the environment in which they would be in. Therefore, the two operations considered are:
\begin{itemize}
    \item A patrolling operation, covering a certain area. A depiction of this concept can be seen in Figure~\ref{fig:expert_soldier}.
    \item Drone formations operations both following a certain path or a target, for example, following the operator.
\end{itemize}

For these use cases, a multi-agent drone system will be developed, as well as a specialized interface, which will enable the expert user to oversee a fleet of drones. These operators will intervene only when necessary, particularly during critical maneuvers or malfunctions, ensuring efficient management of multiple drones simultaneously. 

Within the realm of multi-agent drone operations, pilots face the challenge of various sources of error that could jeopardize mission success and safety. Notably, timing constraints will most likely affect the control of each of the platforms, for example, if they do not receive information from the other agents, they could collide. In order to work out this problem, backup routines~\cite{khayatian2022plan} will be implemented to address timing failures, ensuring continued operation and system safety. This could include the possibility of changing drone trajectory to avoid this issue, or even changing tasks if necessary.

Additionally, the algorithms to solve the routing problems of the different agents are also a source of concern. In~\cite{leon2022multi} they observed that, as the route became larger and more platforms were involved, the results worsened. Although they argue that giving more computation time to the system could improve the output, some applications may not have this extra time to spare when making routing decisions. If the algorithm is centralized, it will definitively be one of the main issues to take into account. Another way to tackle this problem could be to decentralize the decision-making to all agents, for example using a consensus-based control approach~\cite{lizzio2022review}. This can offer the possibility of using different types of platforms in the same mission, as well as customizing the control approach according to the payloads and battery usage of each of the robots~\cite{leon2022multi}. 


Therefore, key functionalities will encompass take-off and landing procedures, communication between the different UASs, sense and avoid capabilities, as well as multi-agent capabilities. 


\begin{table*}[bth]
\centering
\resizebox{0.93\textwidth}{!}{%
\begin{tabular}{|c|cl|}
\hline
\textbf{Module}                                                                                          & \multicolumn{2}{c|}{\textbf{Functionality}}                                                                                                                                                                           \\ \hline
\multirow{12}{*}{\textbf{\begin{tabular}[c]{@{}c@{}}Human-Machine\\ Interface Design\end{tabular}}}      & \multicolumn{1}{c|}{\multirow{4}{*}{\textbf{\begin{tabular}[c]{@{}c@{}}Audio Command \\ and Feedback\end{tabular}}}} & Drone and mission status updates.                                                              \\ \cline{3-3} 
                                                                                                         & \multicolumn{1}{c|}{}                                                                                                & Information about environment (e.g. number of detected enemies, localization of target...).    \\ \cline{3-3} 
                                                                                                         & \multicolumn{1}{c|}{}                                                                                                & Implementation of main drone commands (e.g. take off, land...) and go to waypoint command.     \\ \cline{3-3} 
                                                                                                         & \multicolumn{1}{c|}{}                                                                                                & Implementation of sending pre-programmed missions (e.g. coverage of a certain area).           \\ \cline{2-3} 
                                                                                                         & \multicolumn{1}{c|}{\textbf{Hand Gestures}}                                                                          & Implementation of main drone commands (e.g. take off, land, RPY and thrust).                   \\ \cline{2-3} 
                                                                                                         & \multicolumn{1}{c|}{\multirow{3}{*}{\textbf{Tablet Control}}}                                                        & Visualization of typical drone status information (e.g. IMU, odometry, position, battery...).  \\ \cline{3-3} 
                                                                                                         & \multicolumn{1}{c|}{}                                                                                                & Map visualization and trajectory designer with waypoints.                                      \\ \cline{3-3} 
                                                                                                         & \multicolumn{1}{c|}{}                                                                                                & Mission planner.                                                                               \\ \cline{2-3} 
                                                                                                         & \multicolumn{1}{c|}{\multirow{4}{*}{\textbf{Augmented Reality}}}                                                     & Small interface with basic information about the drone status (e.g. battery status, odometry). \\ \cline{3-3} 
                                                                                                         & \multicolumn{1}{c|}{}                                                                                                & Small interface with map situation of drone.                                                   \\ \cline{3-3} 
                                                                                                         & \multicolumn{1}{c|}{}                                                                                                & Possibility of changing between drone and pilot vision.                                        \\ \cline{3-3} 
                                                                                                         & \multicolumn{1}{c|}{}                                                                                                & Basic control of drone with virtual reality controller.                                        \\ \hline
\multirow{5}{*}{\textbf{Multi-Agent System}}                                                             & \multicolumn{1}{c|}{\multirow{3}{*}{\textbf{Patrolling / Coverage}}}                                                 & Deployment of coverage mission for surveillance purposes.                                      \\ \cline{3-3} 
                                                                                                         & \multicolumn{1}{c|}{}                                                                                                & Possibility of changing drone trajectory for specific task.                                    \\ \cline{3-3} 
                                                                                                         & \multicolumn{1}{c|}{}                                                                                                & Information about environment (e.g. number of detected enemies, localization of target...).    \\ \cline{2-3} 
                                                                                                         & \multicolumn{1}{c|}{\multirow{2}{*}{\textbf{Drone Formations}}}                                                      & Follow trajectory while maintaining formation.                                                 \\ \cline{3-3} 
                                                                                                         & \multicolumn{1}{c|}{}                                                                                                & Follow me function while maintaining formation.                                                \\ \hline
\multirow{3}{*}{\textbf{\begin{tabular}[c]{@{}c@{}}Artificial Intelligence\\ Capabilities\end{tabular}}} & \multicolumn{1}{c|}{\multirow{2}{*}{\textbf{Target Detection}}}                                                      & The system has to detect, localize and track people and vehicles.                              \\ \cline{3-3} 
                                                                                                         & \multicolumn{1}{c|}{}                                                                                                & The system has to differentiate between friendly, unknown and enemy.                           \\ \cline{2-3} 
                                                                                                         & \multicolumn{1}{c|}{\textbf{Defect Detection}}                                                                       & The system has to detect different types of defects.                                           \\ \hline
\end{tabular}
}
\caption{Summary of the different functionalities expected from the HADRON project}
\label{tab-func}
\end{table*}
\subsection{Semi-automated data interpretation} \label{subsec-AI}

In the previous sections, scenarios, and functionalities have been selected which define how the pilots will interact with the drones. The last part of the project consists of the development of assistive tools to mitigate pilots' cognitive burden when handling data streams. The focus of this part of the project will stay on the development of semi-automated methods, where the human remains an important actor in the loop. 

The first tool to be developed will be a computer vision system capable of detecting and localising people and vehicles in various conditions and scenarios, relying on high accuracy and speed. The first challenge that arises is to ensure real-time processing, enabling operators to promptly respond to any detected anomalies. The main goal foreseen for this task is to discern whether the coming person or vehicle is an ally, an enemy, or an unknown. This will allow the operator to focus their attention on the possible threats.

Moreover, the complexity of the task extends to the need for multi-angle detection~\cite{zhang2021research}, accommodating various human postures such as standing, sitting, and lying down. Another challenge will consist of discerning between different types of vehicles, such as cars, buses, or bikes. The system must have the capability to identify subjects even when partially obscured, showcasing its resilience in challenging scenarios where visibility may be compromised.

Furthermore, the requirement for high-speed detection is essential, particularly in environments where vehicles are moving rapidly. Additionally, the system must be orientation-invariant, and capable of detecting vehicles regardless of their positioning to the camera.


The second tool will involve the development of algorithms where a vast array of defects can be identified in different infrastructures~\cite{avdelidis2022defects}. The ensemble of models plays a crucial role in this endeavour, as different types of defects require distinct classes of models built upon each other to ensure comprehensive coverage. Moreover, tracking the evolution of defects throughout their life cycle is essential for effective management and remediation strategies. Achieving an integration among different types of defects is crucial, as the system will be more adaptable to challenging scenarios and environments.


\section{ADVANCED PILOT TRAINING TOOLS} \label{sec-training}
 
Table~\ref{tab-func} summarizes the expected functionalities from the HADRON project. There are three main modules: the human-machine interface design, the multi-agent system, and the artificial intelligence assistive tools. Furthermore, each module is defined as a set of functionalities and technological requirements aimed to be developed by the end of the project.

While making these tools user-friendly, such that they can also be used by operators without extensive training is a primary goal, the operators will still need some form of training to use these tools. This applies also to the multi-agent command and control interface geared towards expert users, as this will induce a whole new control paradigm for these drone operators. Therefore, clear documentation and training tools will be developed for each of the tools developed within this study. 


\section{CONCLUSIONS} \label{sec-conclusions}

This paper outlines the key features developed in the HADRON project. User-friendly drone control interfaces tailored to specific military contexts are defined. The integration of advanced technologies, such as touch screens, gesture-based control, speech recognition, and augmented reality, will create a novel interface that enhances user interaction and situational awareness. A modular system design ensures flexibility, adaptability, and interoperability across various scenarios. Additionally, incorporating vision algorithms and artificial intelligence to automatically detect relevant environmental information will reduce workload to a supervisory role. Overall, this article presents an integrated approach to drone control system development, enhancing operational effectiveness and safety for both novice and expert pilots.


\section*{ACKNOWLEDGMENT}
This project has received research funding from Belgian Defence and the Belgian Royal Higher Institute for Defence in the framework of the HADRON project (DAP22/06) and it is developed in collaboration with Skyebase. 


\bibliographystyle{unsrt}
\bibliography{bibliography}

\end{document}